%% file: main.tex
\documentclass{article}

    \usepackage[preprint,nonatbib]{neurips_2023}

\usepackage[utf8]{inputenc} % allow utf-8 input
\usepackage[T1]{fontenc}    % use 8-bit T1 fonts
\usepackage{hyperref}       % hyperlinks
\usepackage{url}            % simple URL typesetting
\usepackage{booktabs}       % professional-quality tables
\usepackage{amsfonts}       % blackboard math symbols
\usepackage{microtype}      % microtypography
\usepackage{xcolor}         % colors
\usepackage{graphicx}
\usepackage{multicol, multirow, latexsym, amsmath, amssymb}
\usepackage{blindtext}
\usepackage{caption}
\usepackage{subcaption}
\usepackage{tikz}
\usetikzlibrary{positioning}
\usetikzlibrary{shapes}
\usepackage{wrapfig}

\usepackage{bbm}
\usepackage{adjustbox}
\usepackage{verbatimbox}
\usepackage{color, colortbl}
\usepackage{stfloats}

\DeclareMathOperator*{\clip}{clip}
\usepackage{arydshln}
\usepackage{nicefrac}
\usepackage{algorithm, mathtools}
\usepackage{algpseudocode}
\usepackage{tcolorbox}

\title{Scalable Learning of Latent Language Structure \\ With Logical Offline Cycle Consistency}

\author{%
Maxwell Crouse \quad Ramon Astudillo \quad Tahira Naseem \quad Subhajit Chaudhury \\
\textbf{Pavan Kapanipathi} \quad \textbf{Salim Roukos} \quad \textbf{Alexander Gray} \\
\texttt{\{maxwell.crouse, ramon.astudillo, subhajit, alexander.gray\}@ibm.com} \\
\texttt{\{tnaseem, kapanipa, roukos\}@us.ibm.com} \\
IBM Research
}

\begin{document}

\maketitle

\begin{abstract}

% introduce problem and motivation
% Large language models have demonstrated a remarkable ability to scale in effectiveness as the amount of training data they are provided with increases. For many tasks involving natural language, this observed scaling property has pointed to a relatively straightforward means of improving performance: Simply increase the amount of training data given to one's model. Unfortunately, for important tasks that involve expensive annotation costs like semantic parsing (and its converse, text generation), it is challenging to acquire the necessary data at scale. 
% brief description of method
We introduce Logical Offline Cycle Consistency Optimization (LOCCO), a scalable, semi-supervised method for training a neural semantic parser. % for learning to parse from natural language text to semantic, structured representations.
% describe advantages
Conceptually, LOCCO can be viewed as a form of self-learning where the semantic parser being trained is used to generate annotations for unlabeled text that are then used as new supervision. To increase the quality of annotations, our method utilizes a count-based prior over valid formal meaning representations and a cycle-consistency score produced by a neural text generation model as additional signals. Both the prior and semantic parser are updated in an alternate fashion from full passes over the training data, which can be seen as approximating the marginalization of latent structures through stochastic variational inference. The use of a count-based prior, frozen text generation model, and offline annotation process yields an approach with negligible complexity and latency increases as compared to conventional self-learning. As an added bonus, the annotations produced by LOCCO can be trivially repurposed to train a neural text generation model.
% give results pitch
We demonstrate the utility of LOCCO on the well-known WebNLG benchmark where we obtain an improvement of $2$ points against a self-learning parser under equivalent conditions, an improvement of $1.3$ points against the previous state-of-the-art parser, and competitive text generation performance in terms of BLEU score.

\end{abstract}

\input{content/introduction.tex}
\input{content/background.tex}

\input{content/model.tex}

\input{content/experiments.tex}

\input{content/results.tex}

\input{content/conclusions.tex}

\bibliographystyle{abbrv}
\bibliography{paper}

\appendix

\input{content/appendix.tex}

\end{document}

%% file: content/introduction.tex
\section{Introduction}
\label{sec:intro}
% Describe problem

% Large language models (LLMs) have brought about a dramatic shift in how semantic parsers are designed. Until recently, the most successful approaches for semantic parsing (i.e., methods that translate from natural language text to logical forms) were largely hand-crafted and tailored to specific tasks or formalisms \cite{jia-liang-2016-data,dong2018coarse,naseem-etal-2019-rewarding}. While effective within the domains for which they were designed, they were not easily transferable and very laborious to construct (often requiring significant domain-specific expertise). LLMs (e.g., T5 \cite{raffel2020exploring} and BART \cite{lewis2020bart}) have changed this, allowing for more performant systems that require significantly less effort to adapt from one domain to the next.

% However, while LLMs have achieved impressive gains for semantic parsing-related tasks, they still face numerous challenges. First and foremost, LLMs are originally trained for text-only, sequence-to-sequence problems. In contrast, semantic parsing is inherently a text-to-structure problem, wherein the objective is to take in text as input and produce a logical form that is most commonly a tree or graph (see Figure \ref{fig:example_rdf} for an example). Beyond the need to account for explicit structure, LLMs must also overcome the paucity of available training examples which typically require costly expert-level knowledge to collect.

Large language models (LLMs) have brought dramatic gains to semantic parsing-related tasks, allowing for more performant systems that require significantly less effort to adapt from one domain to the next. However, while their impact has been undeniable, they still face numerous challenges. First, LLMs are originally trained for text-only, sequence-to-sequence problems. In contrast, semantic parsing is inherently a text-to-structure problem, wherein the objective is to take in text as input and produce a logical form that is most commonly a tree or graph (see Figure \ref{fig:example_rdf} for an example). Beyond the need to account for explicit structure, LLMs must also overcome a paucity of training examples, which generally require costly expert-level knowledge to collect in this space.

To better generalize to formal, structured representations and alleviate data-scarcity concerns, many high performing text-to-structure and structure-to-text models employ a form of bootstrapping. That is, they fine-tune an initial model using whatever supervised data is available and then subsequently use that model to annotate a large amount of unlabeled text to serve as additional training data \cite{konstas2017neural,xu2020improving,lee2020pushing,bevilacqua2021one,zhou2021structure,ribeiro2021investigating,ribeiro2021investigating,lee2021maximum,bai2022graph}. This form of data augmentation is commonly referred to as \textit{self-learning}, with the parsed data being referred to as pseudo-labels or \textit{silver data}.

Unfortunately, using fine-tuned models to generate data is not always straightforward, since, without specific modifications (e.g., \cite{zhou2021structure,crouse2023laziness}) most pretrained neural models do not offer any well-formedness guarantees. While some approaches that are applied to simpler datasets can sidestep this issue by deriving synthetic examples from grammars induced from the supervised data \cite{jia-liang-2016-data,andreas2020good}, such a strategy is untenable in more realistic open-ended domains. In addition to well-formedness concerns, self-learning models also introduce noise in the labels and are known to saturate in performance relatively quickly (only one round of self-learning labeling and training is used in state-of-the-art systems). More elaborate approaches leveraging latent variable models \cite{yin-etal-2018-structvae} are more robust to such noise and can improve silver data quality over multiple update rounds; however, they require marginalizing over latent discrete structures, which adds significant complexity and computational overhead.

% 
%The silver-data then serves as additional supervision for a new model to be trained with. Unfortunately, while the use of silver data has frequently been shown to improve performance of semantic parsing and text generation methods \cite{,ribeiro-etal-2021-smelting}, it is not a panacea. Often, the predictions of an off-the-shelf model are too noisy for self-learning. Improvements also quickly saturate and most approaches are limited to a single round of self-learning. 

% Broadly speaking, in semantic parsing there are two directions of research that are typically explored to facilitate LLMs in generalizing to data-scarce semantic parsing-related tasks. The first direction adapts the architecture of a model to account for graph structure on the input or output sides \cite{zhou2021structure,zhao-etal-2020-bridging}. The second direction leaves the model as is and instead augments the dataset with enough examples for the model to generalize to structured data \cite{guo2020cyclegt,blinov-2020-semantic,montella-etal-2020-denoising}.

% Describe most common strategy for problem and describe flaw with common strategy

%. For open domain datasets, additional data is taken from a similar, text-only corpus and passed to an existing parsing model to be annotated \cite{guo2020cyclegt,agarwal-etal-2021-knowledge}

\input{content/figure1}

% \begin{figure}[t]
% \centering
% \includegraphics[width=\textwidth]{figures/KG_subset.png}
% \caption{Text-to-RDF example from the WebNLG dataset}
% \label{fig:example_rdf}
% \end{figure}

% Introduce our method and describe how our method addresses issues with previous methods

%{\color{red}Pavan: Connect to the challenges mentioned in the previous paragraph. }
In this work, we introduce Logical Offline Cycle Consistency Optimization (LOCCO), a novel semi-supervised method for training a semantic parser that is designed to address the aforementioned issues. Our method predicts parses for a corpus of text; however, rather than treating the predictions as gold data, each prediction is weighted as a function of two scores: 1) an LLM-produced cycle-consistency score that provides a strong signal as to how faithful a predicted sample is to its original text and 2) a count-based prior probability that gives higher scores to parses that are syntactically valid and share common substructure with other sampled parses across the corpus. The result is a model that is incentivized to produce less-noisy parses that are both coherent with respect to the input text and structurally regular. LOCCO has a principled theoretical foundation as stochastic variational inference \cite{hoffman2013stochastic} and can also be related to offline reinforcement learning. Importantly, our method is straightforward to implement, trivial to parallelize, and comes with very little added computational cost to standard silver data training. In addition to producing a strong semantic parser, the output annotations produced by LOCCO can also be used to train a structure-to-text model.

% State claims

\textbf{Contributions:} (a) We introduce LOCCO, a semi-supervised method for training a neural semantic parser. (b) We demonstrate how the weakly-supervised output of LOCCO can be repurposed to train a strong text generation model. (c) We demonstrate the effectiveness of LOCCO on the well-known WebNLG~2020 \cite{castro-ferreira-etal-2020-2020} benchmark, where we improve semantic parsing by $1.3$ points over the previous state-of-the-art parser while also achieving competitive text generation performance. (d) We compare LOCCO to similar semi-supervised models on the standard ATIS semantic parsing benchmark and demonstrate competitive performance without the need for expensive online sampling. (e) We perform an ablation analysis to determine how each component of LOCCO contributes to overall performance.

%{\color{red} The contributions order (a) (c) (b, d, e) as one for evaluation. Also one this missing in the introduction is p(z) --> I dont see that emphasized enough. Is this on purpose?}

%% file: content/figure1.tex
%\begin{wrapfigure}{l}{0.55\textwidth}
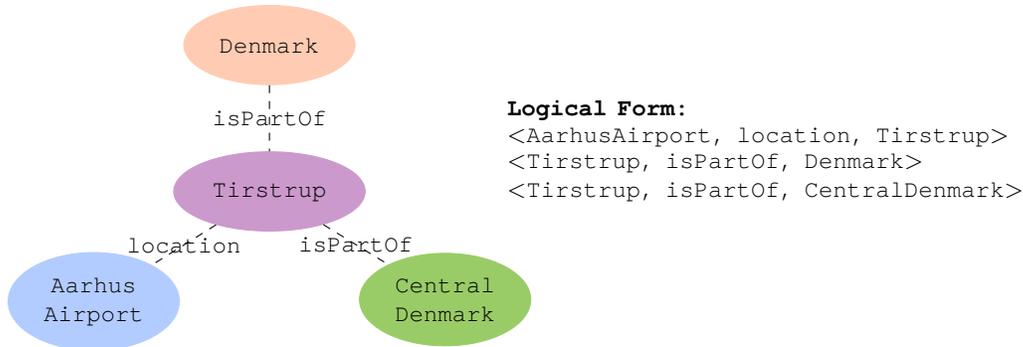
\begin{figure*}
\begin{tikzpicture}[scale=0.5]
\fontfamily{qcr}\selectfont

\definecolor{gblue}{rgb}{0.7, 0.8, 1}
\definecolor{purple}{rgb}{0.8, 0.6, 0.8}
\definecolor{peach}{rgb}{1, 0.8, 0.7}
\definecolor{green}{rgb}{0.6, 0.8, 0.4}

\def \x {0}
\def \y {0}

\node [draw,color=peach,fill=peach,text=black,ellipse,minimum height = 30pt,align=center] (n1) at (\x+0pt,\y+0pt) {\footnotesize Denmark};
%\node [draw,color=purple,fill=purple,text=black,ellipse, below = 50pt of n1,minimum height = 30pt,align=center] (n2)  {\footnotesize Tirstrup};
%\node [draw,color=gblue,fill=gblue,text=black,ellipse, below left = 55pt of n2,minimum height = 35pt,align=center] (n3)  {\footnotesize Aarhus \\ \footnotesize Airport};
%\node [draw,color=green,fill=green,text=black,ellipse, below right = 55pt of n2,minimum height = 35pt,align=center] (n4)  {\footnotesize Central \\ \footnotesize Denmark};
\node [draw,color=purple,fill=purple,text=black,ellipse, below = 25pt of n1,minimum height = 30pt,align=center] (n2)  {\footnotesize Tirstrup};
\node [draw,color=gblue,fill=gblue,text=black,ellipse, below left = 25pt of n2,minimum height = 35pt,align=center] (n3)  {\footnotesize Aarhus \\ \footnotesize Airport};
\node [draw,color=green,fill=green,text=black,ellipse, below right = 25pt of n2,minimum height = 35pt,align=center] (n4)  {\footnotesize Central \\ \footnotesize Denmark};

% \node[fill=white,color=white,text=black, text width = 215pt, align=left](text) at (\x+260pt,\y-45pt) {\footnotesize \textbf{Text:} \\ Aarhus Airport is in Tristrup, Denmark which is part of the Central Denmark Region. \\ \hfill  \\ \textbf{Structure:} \\ $<$AarhusAirport, location, Tirstrup$>$ \\  $<$Tirstrup, isPartOf, Denmark$>$ \\  $<$Tirstrup, isPartOf, CentralDenmark$>$ };

\node[fill=white,color=white,text=black, text width = 200pt, align=left](text) at (\x+380pt,\y-80pt) {\footnotesize \textbf{Logical Form:} \\ $<$AarhusAirport, location, Tirstrup$>$ \\  $<$Tirstrup, isPartOf, Denmark$>$ \\  $<$Tirstrup, isPartOf, CentralDenmark$>$ };

% \node[fill=white,color=white,text=black, text width = 225pt, align=left](text) at (\x+0pt,\y-300pt) {\textbf{Structure:} \\ $<$AarhusAirport, location, Tirstrup$>$ \\  $<$Tirstrup, isPartOf, Denmark$>$ \\  $<$Tirstrup, isPartOf, CentralDenmark$>$ };

%\draw[thick] (n2) -- node[above, sloped]{\footnotesize location} (n3);
%\draw[thick] (n2) -- node[above, sloped]{\footnotesize isPartOf} (n4);
%\draw[thick] (n1) -- node[above, rotate=90]{\footnotesize isPartOf} (n2);

\draw[dashed] (n2) -- node[]{\footnotesize location} (n3);
\draw[dashed] (n2) -- node[]{\footnotesize isPartOf} (n4);
\draw[dashed] (n1) -- node[]{\footnotesize isPartOf} (n2);

\end{tikzpicture}
\caption{Text-to-RDF example from the WebNLG dataset for the sentence, "Aarhus Airport is in Tristrup, Denmark which is part of the Central Denmark Region."}
\label{fig:example_rdf}
\end{figure*}
%\end{wrapfigure}

%% file: content/background.tex
\section{Related Work}

\subsection{Cycle Consistency and Latent Variable Optimization}

End-to-end differentiable Cycle-Consistency (CC) losses concern two probabilistic models relating two domains $p(x \mid z)$ and $p(z \mid x)$, e.g., text / image or text / text. The parameters of both distributions are learned end-to-end via gradient descent to maximize

\begin{equation*}
\mathbb{E}_{p(z \mid x)}[p(x \mid z)] = \int_{z \in D_z} p(x \mid z) p(z \mid x) dz \quad \mbox{ or } \quad \mathbb{E}_{p(z \mid x)}[p(x \mid z)] = \sum_{z \in D_z} p(x \mid z) p(z \mid x) 
\end{equation*}

for continuous or discrete bottleneck variables, respectively. Approaches either optimize for one bottleneck, i.e., $z$, or both $x$ and $z$ simultaneously. CC losses are often used in a semi-supervised fashion by combining datasets where only $z$ or $x$ are available with datasets where they are both available. CC has been shown to be successful in many areas of application, including image transformation \cite{CycleGAN2017}, machine translation \cite{he2016dual,cheng-etal-2016-semi}, speech-to-text, and text-to-speech \cite{hori2019cycle,tjandra2019end}. For all of these domains, the expectations over the output sets are intractable. For continuous domains such as image or speech, it is possible to backpropagate directly through either reparametrization or by collapsing the distribution over the mean\footnotemark\footnotetext{Although not explicitly stated, the output of the composed networks can be interpreted as the mean of a constant variance Laplace distribution, reducing to $|| x - \mathbb{E}_{p(x \mid \mathbb{E} [ z \mid x ])}[x]||_1$} \cite{CycleGAN2017}. For discrete domains, such as text or formal languages, this is not possible and approximations are needed like strong independence assumptions, straight-through approximations \cite{bengio2013estimating, jang2016categorical} as in \cite{tjandra2019end}, the score-function estimator (i.e., REINFORCE \cite{williams1992simple}) used in \cite{he2016dual, hori2019cycle}, or collapsing the distribution to $K$-best \cite{cheng-etal-2016-semi}.

CC losses are related as well to semi-supervised end-to-end learning with latent variables when those variables correspond to interpretable domains, e.g., latent summarization models \cite{miao-blunsom-2016-language}, trees \cite{corro2018differentiable,yin-etal-2018-structvae} and sequence labeling \cite{zhang2017semi}. Most approaches leverage amortized variational inference in the form of Variational Autoencoders \cite{kingma2013auto} and some modified Expectation Maximization \cite{zhang2017semi}. They are restricted to particular structures (e.g. trees) and some require strong independence assumptions \cite{zhang2017semi}. Here, we propose an offline version of variational inference without structure restrictions, that can learn a prior over the latent even when gradient learning is not possible (e.g., rule learning). We also integrate and outperform LLM approaches, which have generally displaced latent variable models.

\subsection{Semantic Parsing and Text Generation}

This work focuses on translating between natural language and formal language domains (see \cite{han2020survey} for a recent survey), e.g., parsing between text and a knowledge graph expressing the semantics of the sentence (as in Figure \ref{fig:example_rdf}). In the area of parsing, there is a large corpus of literature in parse induction \cite{han2020survey} which often involves marginalization over latent structures. Although related to the presented work, these works have two fundamental differences. They are focused on the unsupervised case \cite{clark2010unsupervised} with few works considering semi-supervised learning. They often require strong independence assumptions, e.g., context-free grammars. %Some recent semi-supervised works leverage include \cite{yin-etal-2018-structvae,corro2018differentiable} which are restricted to trees. One fundamental problem with these approaches is that they can not easily leverage pretrained sequence-to-sequence models, that may not be amenable to the independence assumptions necessary.
Beyond parsing, there are a large number of works focused on joint learning of semantic parsing and text generation \cite{guo2020P2,agarwal-etal-2021-knowledge, dognin2021regen, guo2020cyclegt}.
Similar to our work is CycleGT \cite{guo2020cyclegt}, which learns using a CC loss based on iterative back-translation \cite{hoang-etal-2018-iterative}. Also relevant to our work is that of \cite{dognin2021regen}, which jointly learns both transformations without a CC loss, instead applying REINFORCE to approximate non-differentiable losses such as BLEU and METEOR.

%% file: content/model.tex
\section{Our Technique}

\subsection{Desiderata}

The objective of this work is to provide an algorithm for parsing between text, $x$, and formal structured representations, $z$ (i.e., text-to-structure and structure-to-text). The method should be able to harness recent developments in neural network pretraining, as well as available inductive biases and learning algorithms in the formal domain. In short, we aim to

\begin{itemize}
\item be able to leverage strong pretrained transformer models (e.g., BART \cite{lewis2020bart} or T5 \cite{raffel2020exploring}) to learn functions mapping $x \rightarrow z$ and $z \rightarrow x$
\item be able to scale training to large data sizes, which implies overcoming the lack of paired $(x, z)$ data samples
\item be able to incorporate arbitrary constraints into the formal domain $D_z$, which may not be amenable to gradient-based learning, and further update these during training
\end{itemize}

%{\color{red} Pavan: The three points seems very reasonable. The connections to the challenges and contributions should be made here again.}
%We expect the combination of a large pretrained model and logic-based constraints will make the model more robust to noise, 
For this we propose a simple semi-supervised algorithm, inspired by Stochastic Variational Inference \cite{hoffman2013stochastic}, that fulfills the desiderata above. The algorithm reduces to conventional cycle-consistency or self-learning under some simplifications but outperforms both algorithms under the same experimental conditions.

%\subsection{Semi-Amortized Stochastic Variational Inference}

\subsection{Logical Offline Cycle Consistency Optimization}

To begin, we assume access to some supervised data consisting of pairs of plain text $x$ and formal, structured representations $z$, i.e., $(x, z) \in \mathcal{D}^S$. In addition, we also assume access to much larger quantities of only text, i.e., $x \in \mathcal{D}^U$. We start from a probability distribution over sentences that arises from marginalizing over the space of all latent structures $D_z$, e.g., all knowledge-graphs.

\begin{equation}
p(x; \theta) = \sum_{z \in D_z} p(x, z; \theta) 
\end{equation}

Following the usual variational formulation \cite{wainwright2008graphical}, one can express this marginalization in terms of the Evidence Lower Bound (ELBO) and reformulate it in a way that resembles a cycle consistency loss

\begin{align}
\log p(x; \theta) &\geq  \overbrace{\log p(x; \theta) - \mathrm{KL}(q(z \mid x; \phi) \mid\mid p(z \mid x; \theta))}^{\mathrm{ELBO}} \label{variational}\\
&=  \mathbb{E}_{z \sim q(z \mid x; \phi)} [ \log p(x \mid z; \theta) ] - \mathrm{KL}(q(z \mid x; \phi) \mid\mid p(z; \theta))\label{vae}\\
& = \mathbb{E}_{\scriptsize \underbrace{z\!\sim\!q(z\!\mid\!x;\!\phi)}_{\mbox{text-to-structure}}} [ \underbrace{\log p(x \mid z; \theta)}_{\mbox{\footnotesize structure-to-text}} + \underbrace{\log p(z; \theta)}_{\mbox{\footnotesize reasoner}} ] \quad + \underbrace{\mathrm{H}(q_\phi)}_{\mbox{\footnotesize encoding entropy}}\label{em}
\end{align}

where $\mathrm{KL}()$ is the Kullback-Leibler divergence and $\mathrm{H}()$ the entropy. Variational methods alternate between maximizing the ELBO with respect to $\phi$, bringing it closer to the marginal log-likelihood for current $\theta^i$, and maximizing it with respect to $\theta^i$. From Eq.~\ref{variational} one can see that setting $q_\phi$ equal to the posterior  $p(z \mid x; \theta)$ will make the bound tight yielding Expectation Maximization \cite{minka1998expectation}. In this context $q_\phi$ is an auxiliary distribution that is recomputed for each update of $\theta$.

With neural networks the alternate optimization of $\phi$ and $\theta$ with gradient ascent becomes costly. Stochastic Variational Inference (SVI) \cite{hoffman2013stochastic} alleviates this with updates based on a subset of the data, but requires a large number of optimization steps and presents optimization problems \cite{kim2018semi}. Amortized variational inference, best exemplified by Variational Autoencoders (VAEs) \cite{kingma2013auto}, solves this problem by reusing $q_\phi$ across all steps of optimization of $\theta$ and simultaneously updating $\theta$ and $\phi$ via gradient ascent of Eq.~\ref{vae}. VAEs set a parameter-less prior $p(z)$ and do not update it during training.

The approach proposed here takes the formulation in Eq.~\ref{em} and the following design choices

\begin{itemize} 

\item $q(z \mid x; \phi)$ is parametrized by a large language model with pretrained parameters $\Omega$ that maps natural language to formal descriptions, i.e., a semantic parser

\item $p(x \mid z; \rho)$ is parametrized  with a separate copy of $\Omega$. It acts as a conditional language model and is frozen after initialization to prevent adaptation to faulty structures (note here that $\theta$ has been replaced with $\rho$ to reflect separate parameters)

\item $p(z; \theta)$ is a count-based model factorizing the space of possible substructures (e.g., into edges). It incorporates prior knowledge about the formal language, such as valid statements

\item as an initial step, all models $q(z \mid x; \phi)$, $p(x \mid z; \rho)$ and $p(z; \theta)$ are fine-tuned or trained with the labeled dataset $\mathcal{D}^S$ of $(x, z)$ pairs

\item as in SVI we then alternate optimizing $\phi$ and $\theta$, but on \textit{full passes} over the unlabeled $\mathcal{D}^U$. We also use a counts estimator for $\theta$, not gradient, and add $\mathcal{D}^S$ for regularization

\end{itemize} 

As detailed\footnotemark\footnotetext{For ease of explanation, gradient updates shown are just Stochastic Gradient Descent} in Algorithm~\ref{algolocco}, the approach thus combines alternate updates of parameters of SVI, but with full passes over the entire $\mathcal{D}^U \cup \mathcal{D}^S$ with a count-based update. This has both negligible overhead and low variance due to the large amount of samples. Text to structure is a many to one mapping, which makes a count-based model also a good choice i.e. there are fewer labels than for the text counterpart.  With a uniform $p(z; \theta)$, LOCCO reduces to cycle-consistency, albeit with offline updates and frozen conditional language model. With a uniform $p(x \mid z; \rho)$ it reduces to conventional self-learning.

The gradient update of $q(z \mid x; \phi^i)$ includes an expectation over a set $z \in D_z$ that is exponentially large as a function of the input (e.g., graphs) and requires back-propagating through $p(x \mid z; \rho)$ and $p(z; \theta)$. We overcome this with the score function estimator \cite{williams1992simple} which yields following Monte Carlo approximation for the gradient\footnotemark\footnotetext{The entropy term $\mathrm{H}(q_\phi)$ was empirically observed to have no effect and was removed}

\begin{align*}
\nabla_{\phi^i} \mathbb{E}_{z \sim q(z \mid x; \phi^i)}[ V(z, x) ] 
& = \mathbb{E}_{q(z \mid x; \phi^i)} [ \ V(z, x)\nabla_{\phi^i} \log q(z \mid x; \phi^i) \ ]\nonumber\\
& \approx \frac{1}{N} \sum_{n=1}^N V(z_n, x)\nabla_{\phi^i} \log q(z_n \mid x; \phi^i), \quad z_n \sim q(z \mid x; \phi^{i-1})
\end{align*}

\indent where we make the additional \textit{offline} assumption of $\phi^i \approx \phi^{i-1}$ for the purpose of sampling, and

\begin{equation*}
    V(z, x) = \log p(x \mid z; \rho) + \log p(z; \theta^{i-1})
\end{equation*}
\input{content/locco_algo}

This amounts to updating $\phi^i$ with the samples from the previous iteration model $q(z \mid x; \phi^{i-1})$ as if they were gold but weighted by $V(z, x)$ to reflect their possible imperfection. This offline update allows for trivial parallelization of sampling and very delayed communication between the sampler and optimizer, which permits the use of normal disk storage for $V(z, x)$ values (diplayed in Figure~\ref{fig:unsup_train}).

%Conveniently, we only need $V(z, x)$ to compute the gradients for $q(z \mid x; \phi^{i})$. %and thus do not store gradient statistics when we sample from $q(z \mid x; \phi)$. This makes models more memory friendly and opens the door for semi-asynchronous sampling strategies.

% \begin{figure*}
% \centering
% \includegraphics[width=0.85\textwidth]{figures/full_arch.png}
% \caption{Parallel unsupervised training}
% \label{fig:unsup_train}
% \end{figure*}

\input{content/figure2}

The large variance of $V(z, x)$ as an estimate is problematic~\cite{sutton2018reinforcement}, and thus in our implementation we make the following two adjustments from the reinforcement learning literature. First, we normalize the reward as
\begin{equation*}
    A(z, x) = \frac{V(z, x) - \mu}{\sigma}
\end{equation*}
where $\mu$ and $\sigma$ are the mean and standard deviation of the reward across all $N$ samples drawn from $q(z| x; \phi^{i-1})$. Second, following \cite{schulman2017proximal} we substitute $V(z, x)$ by a clipped surrogate objective 
\begin{equation*}
\begin{gathered}
    r_{z_n} = \frac{q(z_n | x; \phi^i)}{q(z_n | x; \phi^{i-1})} \\
    R(z, x)     = \min (r_{z_n} A(z, x), \ \clip(r_{z_n}, 1 - \epsilon, 1 + \epsilon) \ A(z, x))
\end{gathered}
\end{equation*}
where $\epsilon$ is a small constant ($\epsilon = 0.2$ in our experiments). This clipped objective limits the change to $q(z | x; \phi)$ at each training iteration, thus helping to avoid catastrophic forgetting.

The optimization of $\theta$ is carried out with a simple count-based maximum likelihood estimator with smoothing factor $\tau$ and a strong factorization into parts, e.g., subexpressions

\begin{equation*}
    p(z; \theta) = \prod_{s \in \mathrm{parts}(z)} p(s; \theta)  \quad \mbox{with} \quad p(s; \theta) = \theta_s = \frac{\Theta_s}{\sum_{s^\prime \in \mathcal{D}_S}\Theta_{s^\prime}}
\end{equation*}

$s \in \mathrm{parts}(z)$ are all subtrees of the input logical form, e.g., when the target forms are sets of triples (as in WebNLG) a subtree corresponds to an individual triple. $\Theta_s$ contains a count of the number of times part $s$ was observed in the entire corpus and is initialized with $\tau$. $\mathcal{D}_S$ is the set of all data types.

%% file: content/locco_algo.tex
\begin{algorithm}[tb]
   \caption{Logical Offline Cycle Consistency Optimization}
\begin{algorithmic}
\Procedure{LOCCO}{$\mathcal{D}^S,\mathcal{D}^U, q(z \mid x; \phi), p(x \mid z; \rho),p(z; \theta), \Omega, K$}
% \State {\bfseries Input:} Labeled $(x, z)$ dataset $\mathcal{D}^S$ unlabeled $x$ dataset $\mathcal{D}^U$, 
%  pre-trained sequence to sequence parameters $\Omega$, inference network (neural parser) $q(z \mid x; \phi)$, neural formal-to-text $p(x \mid z; \rho)$ counts formal prior $p(z; \theta)$, factorizing criteria $\mathrm{parts}(z)$ into $p \in \mathcal{P}$ possible part types, $\Theta_p$ observed part counts, smoothing factor $\tau$. Semi-supervised epochs $K$ 
% \State {\bfseries Output:} Trained parser parameters $\phi^I$, trained formal prior parameters $\theta^I$, silver dataset $\mathcal{R}^I$
\State $\hphantom{\quad}\mathllap{\rho} \gets \Omega$ \Comment{Initialization}
\State $\hphantom{\quad}\mathllap{\phi^0} \gets \Omega$

\For{batch $B$ sampled without replacement from $\mathcal{D}^S$}  \Comment{Supervised Warm-up}
    \State $\hphantom{\quad}\mathllap{\rho} \gets \rho + \eta \cdot\frac{1}{|B|} \sum_{(x, z) \in B} \nabla_{\rho} \log p(x \mid z; \rho_{k})$
    \State $\hphantom{\quad}\mathllap{\phi^0} \gets \phi^0 +  \eta\cdot\frac{1}{|B|} \sum_{(x, z) \in B} \nabla_{\phi^0} \log q(z \mid x; \phi^0)$
\EndFor

\For{$(x, z) \in \mathcal{D}^S$} \Comment{Parts counts $\Theta_s^0$ initially set to 0}
    \State $\Theta_s^0 \gets \Theta_s^0 + 1$ \quad for $s \in \mathrm{parts}(z)$
\EndFor
    
\State $\theta_s^0 = \frac{\Theta_s^0}{\sum_{s^\prime}\Theta_{s^\prime}^0}$

% SEMI SUPERVISED TRAINING
\For{CC iteration $i \in [1, K]$} \Comment{Semi-supervised Training}
    \State $\mathcal{R}_i \gets \emptyset$ \Comment{Offline Inference Network Update}
    \For{$x \in \mathcal{D}^U$} 
        \For{$z_j \sim q(z \mid x; \phi^{i-1})$} \Comment{Draw $N$ samples}
            \State $v_j = \log p(x \mid z_j; \rho) + \log p(z_j; \theta^{i-1})$
            \State $\mathcal{R}_i \gets \mathcal{R}_i \cup \{ \ (x, z_j, v_j) \ \}$
            \State $\Theta_s^i \gets \Theta_s^i + 1$ \quad for $s \in \mathrm{parts}(z_j)$
            \EndFor
    \EndFor
    \For{batch $B$ sampled without replacement from $\mathcal{D}^U \cup \mathcal{D}^S$} 
        \State $\phi^i \gets \phi^i + \eta\cdot \frac{1}{|B||N|} \sum_{x \in B} \sum_{(x, z_n, R_n) \in \mathcal{R}_i} R_n \nabla_{\phi^i}\log q(z_n \mid x; \phi^i)$
    \EndFor
    
    \State $\theta_s^i = \frac{\Theta_s^i}{\sum_{s^\prime}\Theta_{s^\prime}^i}$ \Comment{Logic Prior Update}

\EndFor
\EndProcedure
\end{algorithmic}
\label{algolocco}
\end{algorithm}

%% file: content/figure2.tex
\begin{figure*}

\begin{tikzpicture}[scale=0.78]
\definecolor{text}{rgb}{0.7, 0.8, 1}
\definecolor{logic}{rgb}{1, 0.8, 0.7}
\definecolor{parser}{rgb}{0.6, 0.8, 0.4}

% ######Left Panel#########

\def \x {50}
\def \y {0}
\node [draw,white,text=black] (title1) at (\x+0pt,\y+0pt) {\textbf{Parallelizable Scoring}};

\def \xt {\x-50}
\def \yt {\y-30}
%stacked rects
\node [draw,white,text=black] (cond) at (\xt+0pt,\yt+0pt) {$\log p(x|z)$};
\node [draw,rounded corners, minimum width =70pt, minimum height =20pt,fill=text,below = 8pt of cond] (text1) {};
\node [draw,rounded corners, minimum width =70pt, minimum height =20pt,fill=text,below right = 4pt of text1.north west] (text2) {};
\node [draw,rounded corners, minimum width =70pt, minimum height =20pt,fill=text,below right = 4pt of text2.north west] (text3) {Text Generator};

\def \xl {\x+50}
\def \yl {\yt}
%stacked rects
\node [draw,white,text=black] (pz) at (\xl+0pt,\yl+0pt) {$\log p(z)$};
\node [draw,rounded corners, minimum width =70pt, minimum height =20pt,fill=logic,below = 8pt of pz] (logic1)  {};
\node [draw,rounded corners, minimum width =70pt, minimum height =20pt,fill=logic, below right = 4pt of logic1.north west] (logic2)  {};
\node [draw,rounded corners, minimum width =70pt, minimum height =20pt,fill=logic, below right = 4pt of logic2.north west] (logic3) {Logic Model};

\node [draw,white,text=black] (z) at (\x+5pt,\y-120pt) {$z$ - Structure (Sampled)};

\def \xs {\x}
\def \ys {\y-150}
%stacked rects
% \node [draw,white,text=black] (title1) at (\x+0pt,\y+0pt) {$\log p(z)$};
\node [draw,rounded corners, minimum width =70pt, minimum height =20pt,fill=parser,below = 30pt of z] (parser1) {};
\node [draw,rounded corners, minimum width =70pt, minimum height =20pt,fill=parser,below right = 4pt of parser1.north west] (parser2)  {};
\node [draw,rounded corners, minimum width =70pt, minimum height =20pt,fill=parser,below right = 4pt of parser2.north west] (parser3){Semantic Parser};

\node [draw,white,text=black,below = 40pt of parser1] (x)  {$x$ - Observed Text};

% ######Middle Panel#########

\def \xx {\x+180}
\def \yy {0}
\node [draw,white,text=black] (title2) at (\xx+0pt,\yy+0pt) {\textbf{Update Variational Parameters}};
\node [draw,white,text=black] (joint) at (\xx+0pt,\yy-60pt) {$\log p(x|z)+ \log p(z)$};
\node [draw,white,text=black] (z2) at (\xx+0pt,\yy-90pt) {$z$ - Structure (Sampled)};
\node [draw,rounded corners, minimum width =70pt, minimum height =20pt,fill=parser,below = 30pt of z2] (parser4) {Semantic Parser};
\node [draw,white,text=black,minimum width =100pt,below = 30pt of parser4] (x2) {$x$ - Observed Text};

% ######Right Panel#########
\def \xxx {\xx+150}
\def \yyy {0}
\node [draw,white,text=black] (title2) at (\xxx+0pt,\yyy+0pt) {\textbf{Update Reasoner}};
\node [draw,white,text=black] (z3) at (\xxx+0pt,\yyy-90pt) {$z$ - Structure (Sampled)};
\node [draw,rounded corners, minimum width =70pt, minimum height =20pt,fill=logic, below = 30pt of z3] (logic4)  {Logic Model};
\node [draw,white,minimum width =100pt, below = 30pt of logic4] (x3) {$x$ - Observed Text};

% Links
\draw [-latex,thick,shorten >=10pt] (x) -- (parser1){};
\draw [-latex,thick] (parser1) -- node[right,pos=0.5]{$z \sim p(z|x)$}(z);
\draw [-latex,thick,shorten >=7pt] (z) -- (text3){};
\draw [-latex,thick,shorten >=7pt] (z) -- (logic3){};
\draw [-latex,thick,shorten >=7pt] (x.west) to[out=180,in=270] (text3.south west){};
\draw [-latex,thick,dashed,shorten >=3pt,shorten <=3pt] (pz.east) -- (joint.west){};
\draw [-latex,thick,dashed,shorten >=3pt,shorten <=3pt] (z.east) -- (z2.west){};
\draw [-latex,thick,dashed,shorten >=3pt,shorten <=3pt] (z2.east) -- (z3.west){};

\draw [-latex,thick,shorten >=10pt,shorten <=10pt] (z2.west) -- node[midway,above,rotate=90]{\footnotesize backpropagation}(x2.west){};
\draw [-latex,thick,shorten >=10pt,shorten <=10pt] (z3.west) -- node[midway,above,rotate=90]{\footnotesize update counts}(x3.west){};

\draw [-latex,thick,shorten >=3pt] (x2) -- (parser4){};
\draw [-latex,thick,shorten >=3pt,shorten <=3pt] (parser4) -- (z2){};

\draw [-latex,thick,shorten >=3pt,shorten <=3pt] (logic4) -- (z3){};

\end{tikzpicture}

\caption{Parallelization details for LOCCO semi-supervised training}
\label{fig:unsup_train}

\end{figure*}
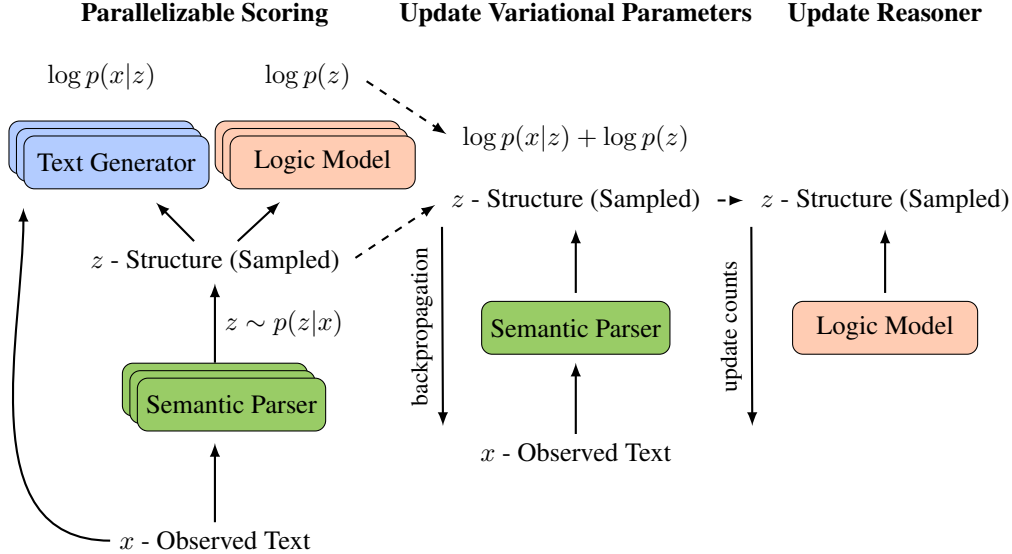

%% file: content/experiments.tex
\section{Experiments}

We performed an extensive series of evaluations utilizing two datasets, the english version of the WebNLG2020+ dataset \cite{castro-ferreira-etal-2020-2020} and the ATIS dataset as processed as in \cite{dong-lapata-2016-language}. Our primary goals were to determine if LOCCO produces an effective semantic parser and to assess the contribution of each component of LOCCO to semantic parsing performance. In addition, we were also interested to learn if the outputs of LOCCO could be used to train a reasonable text generation system. For WebNLG we include a comparison with recent systems in both parsing and generation, including the state-of-the art. We also include a self-learning baseline, component ablation, and investigation into the effect of iterative training. For ATIS we assess the effect of training data size on performance. 

\subsection{Datasets}

WebNLG is a dataset where each example is a pairing of text with a set of RDF triples. The dataset contains 13,211 training pairs, 1,667 validation pairs, 2,155 pairs for testing semantic parsing, and 1,779 pairs for testing text generation. Its use in this work was motivated by it being a well-known, open-domain benchmark with several systems to compare against that tests both semantic parsing and text generation. For our WebNLG experiments, silver data consisted of 50,000 sentences randomly selected from the TekGen corpus \cite{agarwal-etal-2021-knowledge}. TekGen is a large-scale dataset intended to provide a more realistic testbed for knowledge extraction. It is comprised of text instances from Wikipedia that have been annotated with a state-of-the-art RDF triple semantic parser. As our system is intended to operate with unlabeled data, we used \textit{only} the text from examples extracted from the corpus.

ATIS is a semantic parsing dataset where each example is a pairing of text with a $\lambda$-calculus logical form. The dataset consists of 4,434 training pairs, 490 validation pairs, and 447 test pairs. We reproduce the StructVAE experimental setup in \cite{yin-etal-2018-structvae} where the training set is split into two disjoint subsets of varying sizes. One of the subsets is treated as the gold dataset (i.e., keeping both the text and logical form) and the other is considered the silver dataset (i.e., keeping only the text). This both tests LOCCO's performance for different data sizes and shows how the approach generalizes to more complex meaning representations than straightforward RDF-triples. We also provide StructVAE results for completeness\footnotemark\footnotetext{It is important to note that StructVAE preceded the use of LLMs and is thus at a clear disadvantage}. 

We performed minimal processing of both datasets. The parentheses of each logical form were replaced with \texttt{<SE>} and \texttt{</SE>} tags to demarcate expression boundaries, and each text-to-structure and structure-to-text example was prompted with either "Text to Graph:" or "Graph to Text:", respectively. For WebNLG, we applied the following transformations to each example: 1) The subject, relation, and object were marked with \texttt{<S>}, \texttt{<R>}, and \texttt{<O>} tags, respectively and 2) the camel-cased text of each triple element was split into individual words based on capitalization.

For WebNLG, we used the provided evaluation scripts to assess performance. For semantic parsing, there were four types of scored matches; however, for space, we display only the Exact Match metric in our results section (we provide the full table of results in the Appendix). For text generation, we provide results for BLEU, METEOR, and chrF++, with BLEU being our primary metric. With ATIS, we report exact-match accuracy, i.e., whether or not the generated form exactly matched the target.

\subsection{Training Details}

For all experiments, we used pretrained BART-large \cite{lewis2020bart} as our model. The semantic parser was taken to be the model produced at the last iteration of semi-supervised training. For each iteration, we evaluated the model on validation data after every 2500 update steps and kept only the top performing model. We list all hyperparameters in the Appendix.

For our text generation experiments, we aimed to keep the training setup as simple as possible. We first used the final model from text-to-structure training to generate a new set of data (following the same setup as each of the prior iterations). Then, we flipped the generated annotations, converting each pair $(x, z)$ into $(z, x)$. Following the conversion, we trained a BART-large model from scratch on the sampled annotations in the same way as was done for the semantic parsing experiments.

%% file: content/results.tex
{
\tabcolsep=0.1cm
\begin{table}[t]
\begin{center}
\begin{subtable}{0.46\textwidth}
\centering
\footnotesize
\begin{tabular}{lccc}
\toprule
Method &  F1 & Precision & Recall  \\
\midrule
LOCCO (Ours) & 0.736 & 0.729 & 0.749  \\
\hdashline
ReGen  \cite{dognin2021regen} & 0.723 & 0.714 & 0.738  \\
Grapher  \cite{melnyk2021grapher} & 0.709 & 0.702 & 0.720 \\
Amazon AI \cite{guo2020P2} & 0.689 & 0.689 & 0.690 \\
bt5 \cite{agarwal2020machine} & 0.682 & 0.670 & 0.701 \\
CycleGT \cite{guo2020cyclegt} & 0.342 & 0.338 & 0.349 \\
Baseline \cite{castro-ferreira-etal-2020-2020} & 0.158 & 0.154 & 0.164 \\
\bottomrule
\end{tabular}
\caption{Semantic parsing ranked by F1}
\label{res:t2s_results}
\end{subtable}
\hspace{1em}
\begin{subtable}{0.46\textwidth}
\centering
\footnotesize
\begin{tabular}{lccc}
\toprule
Method & BLEU & METEOR & chrF++  \\
\midrule
ReGen  \cite{dognin2021regen} & 0.563 & 0.425 & 0.706 \\
LOCCO (Ours) & 0.552 & 0.406 & 0.691 \\
\hdashline
Amazon AI \cite{guo2020P2} & 0.540 & 0.417 & 0.690 \\
OSU NLG \cite{li2020leveraging} & 0.535 & 0.414 & 0.688  \\
FBConvAI \cite{Yang2020ImprovingTP} & 0.527 & 0.413 & 0.686 \\
bt5 \cite{agarwal2020machine} & 0.517 & 0.411 & 0.679 \\
NUIG-DSI \cite{pasricha-etal-2020-nuig} & 0.517 & 0.403 & 0.669 \\
cuni-ufal \cite{kasner-dusek-2020-train} & 0.503 & 0.398 & 0.666 \\
CycleGT \cite{guo2020cyclegt} & 0.446 & 0.387 & 0.637  \\
Baseline \cite{castro-ferreira-etal-2020-2020} &  0.406 & 0.373 & 0.621  \\
RALI \cite{lapalme2020rdfjsrealb} & 0.403 & 0.386 & 0.634 \\
\bottomrule
\end{tabular}
\caption{Text generation ranked by BLEU}
\label{res:s2t_results}
\end{subtable}
\caption{WebNLG test set results for semantic parsing (F1 Strict) and text generation (BLEU, METEOR, chrF++). Dashed line includes existing results matching or outperfomring LOCCO}
\label{res:both_results}
\end{center}
\end{table}
}

\section{Results}

\subsection{WebNLG}

Our main results can be found in Tables \ref{res:t2s_results} and \ref{res:s2t_results}, which show the performance of our model for both semantic parsing and text generation as compared to other approaches. As can be seen in Table \ref{res:t2s_results}, LOCCO achieves state-of-the-art performance on the semantic parsing task, with a notable improvement (0.13 F1) over the next best model ReGen \cite{dognin2021regen}. Importantly, our model achieves these results without any special modifications to the underlying large language model (e.g., constrained output, triple reordering, etc.) that are common to the other approaches on this dataset \cite{guo2020P2,agarwal2020machine}. 

In Table \ref{res:s2t_results}, we see that our approach yields a reasonably performant text generation system. It outperforms all other approaches (many of which were specifically designed for RDF-to-text) but ReGen. This is significant, as the text generation system we use is functionally a byproduct of our process for producing a semantic parser. It has no tailored architectural features and is simply trained using the data produced by our semantic parser.% This drastically lessens the expert knowledge required to apply our system.

\begin{table*}[t]
\begin{center}
\begin{small}
\begin{tabular}{lccccccc}
\toprule
Method & Reward Function & \multicolumn{3}{c}{Semantic Parsing} & \multicolumn{3}{c}{Text Generation} \\
\cmidrule(lr){3-5}
\cmidrule(lr){6-8}
 &  & F1 & Precision & Recall & BLEU & METEOR & chrF++  \\
\midrule
LOCCO & $\log p(x | z) + \log p(z)$ & 0.736 & 0.729 & 0.749 & 0.552 & 0.406 & 0.691 \\
LOCCO & $\log p(x | z)$ & 0.733 & 0.725 & 0.745 & 0.551 & 0.416 & 0.692 \\
LOCCO & $\log p(z)$ & 0.716 & 0.710 & 0.728 & 0.519 & 0.405 & 0.676 \\
Greedy SL & -- & 0.715 & 0.708 & 0.726 & 0.507 & 0.401 & 0.663 \\
Sampling SL & -- & 0.718 & 0.712 & 0.728 & 0.524 & 0.407 & 0.677 \\
Gold-Only & -- & 0.691 & 0.684 & 0.703 & 0.526 & 0.406 & 0.678 \\
\bottomrule
\end{tabular}
\end{small}
\end{center}
\caption{WebNLG ablation results for semantic parsing (in terms of Exact Match) and text generation}
\label{res:abl_results}
\end{table*}

\subsubsection{Ablation Experiments}

In addition to our main results, we also perform extensive ablation experiments to determine the contributions of each element of our training objective. In Table \ref{res:abl_results} we show various ablations to the reward function of our model, self-learning (SL) where the annotated silver parses are drawn from either greedy or sampling-based decoding, and gold-only training where no silver data is used.

From the table, it can be seen that using silver data in any capacity leads to improved performance over gold-only training. This is a promising result, as it suggests that our approach could be used to improve the other state-of-the-art models that did not train with external data, e.g., \cite{dognin2021regen}. The results indicate that greedy and count-based rewards produce roughly the same performance. This is somewhat unsurprising, as the count-based model should reward higher-probability triples that are sampled frequently (i.e., those that would be produced by greedy decoding). The most important result is that the combination of cycle-consistency and the count-based logic model produces the best performance, better than either score individually.

\begin{figure*}[t]
\centering
\includegraphics[width=\textwidth]{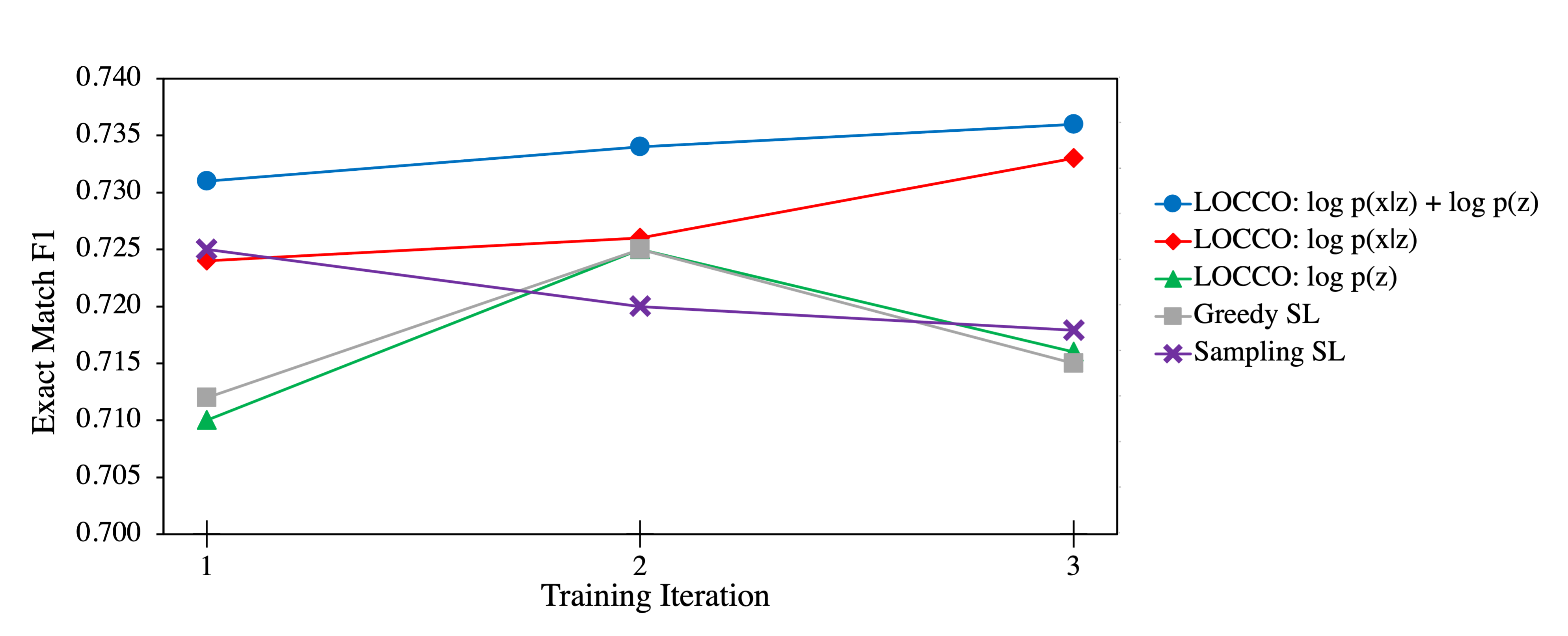}
\caption{Semantic parsing performance across training iterations as measured by Exact Match F1}
\label{fig:across_epochs}
\end{figure*}

\subsubsection{Performance Across Epochs}

Though the main results were based on the model trained at the final iteration, we were also interested in the performance of each model in the intermediate iterations of training. The across-iteration performance is shown in Figure \ref{fig:across_epochs}, where we see that the unablated version of LOCCO demonstrates consistently higher performance than the other versions. In addition, consistent with our remarks in Section \ref{sec:intro}, we see that sampling-based self-learning produces strong results at first but then degenerates over time. Another interesting result is that greedy self-learning is largely equivalent in performance to LOCCO when only the prior $p(z)$ is used for the reward. Again, we suspect that this is due to the nature of our count-based model which upweights logical forms with frequent triples, i.e., those considered more likely by our neural model, and thus more likely to also be a part of the greedy decoding.

\subsection{ATIS}

\begin{table}[t]
\begin{center}
\begin{small}
\begin{tabular}{lccccccc}
\toprule
$|\mathcal{D}^S|$ & \multicolumn{3}{c}{LOCCO} & \multicolumn{3}{c}{StructVAE \cite{yin-etal-2018-structvae}} & SOTA \cite{cao-etal-2019-semantic} \\
\cmidrule(lr){2-4}
\cmidrule(lr){5-7}
& Gold-Only & Self-Learning & $R(z, x)$ & Gold-Only & Self-Learning & $R(z, x)$ & \\
\midrule
500 & 71.9 & 76.8 & 75.9 & 63.2 & 65.3 & 66.0 & -- \\
1000 & 77.0 & 77.9 & 81.0 & 74.6 & 74.2 & 75.7 &  -- \\
2000 & 86.1 & 86.4 & 87.1 & 80.4 & 83.3 & 82.4 &  -- \\
3000 & 85.9 & 87.3 & 87.7 & 82.8 & 83.6 & 83.6 &  -- \\
4434 & 86.3 & -- & -- & 85.3 & -- & -- & 89.1 \\
\bottomrule
\end{tabular}
\end{small}
\end{center}
\caption{Semantic parsing on ATIS for various training set sizes. The last row reflects when all supervised data is used (i.e., there is no additional data for semi-supervised training)}
\label{res:atis_results}
\end{table}

Table \ref{res:atis_results} shows our results on the ATIS semantic parsing dataset as compared to StructVAE \cite{yin-etal-2018-structvae}. The first column shows the size of the gold dataset, $\mathcal{D}^{S}$, while the remaining columns provide results for each system. For both LOCCO and StructVAE we distinguish between training with gold-only (i.e., only examples in $\mathcal{D}^{S}$ used), self-learning, and $R(z, x)$ (i.e., when all silver examples are scored) settings. In addition, we include the current state-of-the-art method \cite{cao-etal-2019-semantic} for reference.

Similar to StructVAE, LOCCO demonstrates performance gains over both the supervised and self-learning settings (with the exception of $|\mathcal{D}^{S}| = 500$). This suggests that, like StructVAE, our approach is producing more meaningful annotations of the unlabeled data than pure self-learning. This is important, as the target meaning representation, $\lambda$-calculus, is significantly more complex than the representations required for WebNLG. Key to note is that our results are achieved with offline sampling and scoring, while theirs requires sampling during training. Lastly, we emphasize that our objective with this experiment was not to compare raw performance, but was instead to determine if our approach yielded similar gains as compared to the supervised and self-learning settings. While our model demonstrated an overall improvement as compared to theirs, this is likely attributable to our much stronger pretrained model (they use an LSTM with GLOVE embeddings \cite{pennington2014glove}) that provided a better baseline performance.

%% file: content/conclusions.tex
\section{Conclusions}

In this paper, we introduced Logical Offline Cycle Consistency Optimization (LOCCO), a novel semi-supervised method for training a neural semantic parser. Our method was inspired by Stochastic Variational Inference, and designed from the ground up to be scalable, take advantage of powerful pretrained LLMs, and be able to incorporate inductive biases relevant to the formal domain. We demonstrated the effectiveness of our model on two standard benchmark datasets, where it achieved strong performance for both semantic parsing and text generation.

%% file: content/appendix.tex
\section{Appendix}

\subsection{Hyperparameters and Hardware Details}

\begin{table*}[h]
\begin{center}
\begin{small}
\begin{tabular}{lcc}
\toprule
Hyperparameter & WebNLG & ATIS \\
\midrule
Dropout & 0.0 & 0.0 \\
Batch size & 8 & 8 \\
Learning rate & 5e-6 & 5e-6 \\
Training iterations & 3 & 3 \\
Training epochs per iteration & 1 & 100 \\
Patience & 5 & 5 \\
Temperature & 1.0 & 1.0 \\
Top-\textit{p} & 0.95 & 0.95 \\
Number of samples \textit{N} & 5 & 5 \\
\bottomrule
\end{tabular}
\end{small}
\end{center}
\caption{Hyperparameters for WebNLG and ATIS}
\label{tbl:hparams}
\end{table*}

For hyperparameter choices, the batch size and learning rate were chosen based on common defaults for BART-large. Sampling-based parameters (i.e., temperature, top-\textit{p}, and number of samples \textit{N}) were similarly chosen to be those commonly found in RL-based works. The decision to disable dropout was to reduce a possible source of randomness in the results after observing it had no effect on validation performance. We chose the number of epochs to run per training iteration after observing that BART overfit the dataset (in the case of WebNLG) or saturated in performance (in the case of ATIS) on validation data around that point.

In terms of hardware, our experimental setup utilized a HPC cluster with CPU and GPU machines running Red Hat Enterprise Linux release 8.7 (Ootpa). CPU machines were used for all non-neural preprocessing and GPU machines were used for model training. Both CPU and GPU machines had 2 CPU cores and 100GB of RAM. GPU machines ran an NVIDIA V100 Tensor Core GPU with 40GB of GPU memory.

\clearpage
\newpage

\subsection{Full Results Tables}

WebNLG measures semantic parsing performance with 4 different metrics, each reflecting a degree of match quality. The 4 metrics are Exact, Entity Type, Partial, and Strict. In Tables \ref{res:full_t2s_results} and \ref{res:full_abl_results}, we show results for all of these metrics (and include the text generation metrics for ease of reference).

\begin{table*}[h]
\begin{center}
\begin{small}
\begin{tabular}{llccc}
\toprule
Method & Match & F1 & Precision & Recall  \\
\midrule
ReGen  \cite{dognin2021regen} & Exact & 0.723 & 0.714 & 0.738  \\
 & Entity Type & 0.807 & 0.791 & 0.835 \\
 & Partial & 0.767 & 0.755 & 0.788 \\
 & Strict & 0.720 & 0.713 & 0.735 \\
\midrule
Grapher  \cite{melnyk2021grapher} & Exact & 0.709 & 0.702 & 0.720 \\
 & Entity Type & -- & -- & -- \\
 & Partial & 0.735 & 0.725 & 0.750 \\
 & Strict & 0.706 & 0.700 & 0.717 \\
\midrule
Amazon AI \cite{guo2020P2} & Exact & 0.689 & 0.689 & 0.690 \\
 & Entity Type & 0.700 &  0.699 &  0.701 \\
 & Partial & 0.696 & 0.696 & 0.698 \\
 & Strict & 0.686 & 0.686 & 0.687 \\
\midrule
bt5 \cite{agarwal2020machine} & Exact & 0.682 & 0.670 & 0.701 \\
 & Entity Type &  0.737 & 0.721 & 0.762 \\
 & Partial &  0.713 & 0.700 & 0.736  \\
 & Strict &  0.675 & 0.663 & 0.695 \\
\midrule
CycleGT \cite{guo2020cyclegt} & Exact & 0.342 & 0.338 & 0.349 \\
 & Entity Type &  0.343 & 0.335 & 0.356 \\
 & Partial & 0.360 & 0.355 & 0.372 \\
 & Strict & 0.309 & 0.306 & 0.315 \\
\midrule
Baseline \cite{castro-ferreira-etal-2020-2020} & Exact & 0.158 & 0.154 & 0.164 \\
 & Entity Type &  0.193 & 0.187 & 0.202 \\
 & Partial & 0.200 & 0.194 & 0.211 \\
 & Strict & 0.127 & 0.125 & 0.130 \\
\midrule
LOCCO (Ours) & Exact & 0.736 & 0.729 & 0.749  \\
 & Entity Type & 0.808 & 0.796 & 0.829 \\
 & Partial & 0.775 & 0.766 & 0.793 \\
 & Strict & 0.733 & 0.726 & 0.745 \\
\bottomrule
\end{tabular}
\end{small}
\end{center}
\caption{Full WebNLG results for semantic parsing}
\label{res:full_t2s_results}
\end{table*}

{
\tabcolsep=0.1cm
\begin{table*}[h]
\begin{center}
\begin{small}
\begin{tabular}{lllcccccc}
\toprule
Method & Reward Function & \multicolumn{4}{c}{Semantic Parsing} & \multicolumn{3}{c}{Text Generation} \\
\cmidrule(lr){3-6}
\cmidrule(lr){7-9}
& & Match & F1 & Precision & Recall & BLEU & METEOR & chrF++  \\
\midrule
LOCCO & $\log p(x | z) + \log p(z)$ & Exact & 0.736 & 0.729 & 0.749 &  0.552 & 0.406 & 0.691 \\
 &  & Entity Type & 0.808 & 0.796 & 0.829 & & & \\
  & & Partial & 0.775 & 0.766 & 0.793 & & & \\
  & & Strict & 0.733 & 0.726 & 0.745 & & & \\
\midrule
LOCCO & $\log p(x | z)$ & Exact & 0.733 & 0.725 & 0.745 & 0.551 & 0.416 & 0.692 \\
&  & Entity Type & 0.804 & 0.791 & 0.825 & & & \\
 & & Partial & 0.771 & 0.761 & 0.788 & & & \\
 & & Strict & 0.729 & 0.722 & 0.742 & & & \\
\midrule
LOCCO & $\log p(z)$ & Exact & 0.716 & 0.710 & 0.728 & 0.519 & 0.405 & 0.676 \\
&  & Entity Type & 0.798 & 0.786 & 0.818 & & & \\
 & & Partial & 0.760 & 0.751 & 0.777 & & & \\
 & & Strict & 0.712 & 0.705 & 0.723 & & & \\
\midrule
Greedy SL & & Exact & 0.715 & 0.708 & 0.726 & 0.507 & 0.401 & 0.663 \\
 & & Entity Type & 0.786 & 0.775 & 0.805 & & & \\
 & & Partial & 0.755 & 0.745 & 0.770 & & & \\
 & & Strict & 0.708 & 0.702 & 0.719 & & & \\
\midrule
Sampling SL & & Exact & 0.718 & 0.712 & 0.728 & 0.524 & 0.407 & 0.677 \\
 & & Entity Type & 0.785 & 0.775 & 0.802 & & & \\
 & & Partial & 0.755 & 0.747 & 0.769 & & & \\
 & & Strict & 0.713 & 0.707 & 0.723 & & & \\
\midrule
Gold-Only & & Exact & 0.691 & 0.684 & 0.703 & 0.526 & 0.406 & 0.678 \\
 & & Entity Type & 0.762 & 0.750 & 0.783 & & & \\
 & & Partial & 0.732 & 0.722 & 0.749 & & & \\
 & & Strict & 0.684 & 0.677 & 0.696 & & & \\
\bottomrule
\end{tabular}
\end{small}
\end{center}
\caption{Full WebNLG ablation results}
\label{res:full_abl_results}
\end{table*}
}

\clearpage
\newpage

\subsection{Dataset Examples}

Here we provide additional examples from both of the datasets used in this paper. In addition, we show how the $\mathrm{parts}(z)$ function breaks down a logical form. Figure \ref{fig:webnlg_examples} provides examples from the WebNLG corpus while Figure \ref{fig:atis_examples} provides examples from the ATIS corpus.

\subsubsection{WebNLG}

\begin{figure*}[h]
\begin{subfigure}{\textwidth}
\footnotesize
\begin{flushleft}
{\texttt{Text \textit{x}: \\ \quad "The Aarhus is the airport of Aarhus, Denmark."
\\ 
Logical Form \textit{z}: \\ \quad (<S> Aarhus Airport <R> city served <O> "Aarhus, Denmark")
\\
\textrm{parts}(z): \\ \quad \{ (<S> Aarhus Airport <R> city served <O> "Aarhus, Denmark") \}
}
}
\end{flushleft}
\end{subfigure}
\vskip 0.25in
\begin{subfigure}{\textwidth}
\footnotesize
\begin{flushleft}
{\texttt{Text \textit{x}: \\ \quad "The Acharya Institute of Technology's campus is located in Soldevanahalli, Acharya Dr. Sarvapalli Radhakrishnan Road, Hessarghatta Main Road, Bangalore, India, 560090. It was established in 2000 and its director is Dr G.P. Prabhukumar. It is affiliated to the Visvesvaraya Technological UNiversity in Belgaum and has 700 postgraduate students."
\\ 
Logical Form \textit{z}: \\ \quad (<S> Acharya Institute of Technology <R> affiliation <O> Visvesvaraya Technological University) \\ \quad (<S> Acharya Institute of Technology <R> campus <O> "In Soldevanahalli, Acharya Dr. Sarvapalli Radhakrishnan Road, Hessarghatta Main Road, Bangalore – 560090.") \\ \quad (<S> Acharya Institute of Technology <R> country <O> "India") \\ \quad (<S> Acharya Institute of Technology <R> established <O> 2000) \\ \quad (<S> Acharya Institute of Technology <R> motto <O> "Nurturing Excellence") \\ \quad (<S> Acharya Institute of Technology <R> state <O> Karnataka) \\ \quad (<S> Visvesvaraya Technological University <R> city <O> Belgaum)
\\
\textrm{parts}(\textit{z}): \\ \quad \{ (<S> Acharya Institute of Technology <R> state <O> Karnataka) \\ \quad \quad (<S> Visvesvaraya Technological University <R> city <O> Belgaum), \ldots \ \}
}
}
\end{flushleft}
\end{subfigure}
\caption{RDF examples from the WebNLG corpus}
\label{fig:webnlg_examples}
\end{figure*}

\subsubsection{ATIS}

\begin{figure*}[h]
\begin{subfigure}{\textwidth}
\footnotesize
\begin{flushleft}
{\texttt{Text \textit{x}: \\ \quad "is there a flight from ci1 to ci0 which connect in ci2"
\\ 
Logical Form \textit{z}: \\ \quad (lambda \$0 e (and (flight \$0) (from \$0 ci1) (to \$0 ci0) (stop \$0 ci2))) 
\\
\textrm{parts}(\textit{z}): \\ \quad \{ (flight \$0), (from \$0 ci1), \ldots \ \}
}
}
\end{flushleft}
\end{subfigure}
\vskip 0.25in
\begin{subfigure}{\textwidth}
\footnotesize
\begin{flushleft}
{\texttt{Text \textit{x}: \\ \quad "where is ap0 locat"
\\ 
Logical Form \textit{z}: \\ \quad (lambda \$0 e (loc:t ap0 \$0)) 
\\
\textrm{parts}(\textit{z}): \\ \quad \{ (loc:t ap0 \$0), (lambda \$0 e (loc:t ap0 \$0)) \}
}
}
\end{flushleft}
\end{subfigure}
\vskip 0.25in
\begin{subfigure}{\textwidth}
\footnotesize
\begin{flushleft}
{\texttt{Text \textit{x}: \\ \quad "show me the fare from ci1 to ci0"
\\ 
Logical Form \textit{z}: \\ \quad (lambda \$0 e (exists \$1 (and (from \$1 ci1) (to \$1 ci0) (= (fare \$1) \$0))))
\\
\textrm{parts}(\textit{z}): \\ \quad \{ (= (fare \$1) \$0), (fare \$1), \ldots \ \}
}
}
\end{flushleft}
\end{subfigure}
\caption{$\lambda$-calculus examples from the ATIS corpus}
\label{fig:atis_examples}
\end{figure*}